\documentclass[10pt,english,journal]{IEEEtran}
\usepackage[latin9]{inputenc}
\usepackage{color}
\usepackage{babel}
\usepackage{array}
\usepackage{textcomp}
\usepackage{url}
\usepackage{multirow}
\usepackage{amsmath}
\usepackage{amssymb}
\usepackage{stmaryrd}
\usepackage{graphicx}
\usepackage[unicode=true,
 bookmarks=false,
 breaklinks=false,pdfborder={0 0 1},backref=section,colorlinks=false]
 {hyperref}

\makeatletter

\providecommand{\tabularnewline}{\\}

\usepackage{babel}
\usepackage{babel}
\usepackage{babel}
\usepackage{algorithm}
\usepackage{tabularx}\usepackage{algpseudocode}

\newcommand{\multiline}[1]{%
\begin{tabularx}{\dimexpr\linewidth-\ALG@thistlm}[t]{@{}X@{}}
#1
\end{tabularx}
}

\newcommand{\lyxmathsym}[1]{\ifmmode\begingroup\def\b@ld{bold}
\text{\ifx\math@version\b@ld\bfseries\fi#1}\endgroup\else#1\fi}

\providecommand{\tabularnewline}{\\}
\floatstyle{ruled}
\newfloat{algorithm}{tbp}{loa}
\providecommand{\algorithmname}{Algorithm}
\floatname{algorithm}{\protect\algorithmname}

\providecommand{\tabularnewline}{\\}
\floatstyle{ruled}
\newfloat{algorithm}{tbp}{loa}
\providecommand{\algorithmname}{Algorithm}
\floatname{algorithm}{\protect\algorithmname}

\usepackage{babel}

\usepackage{url}

\usepackage{array}
\usepackage{multirow}

\makeatother

\begin{document}

\title{Multi-level Semantic Feature Augmentation for One-shot Learning}

\author{Zitian Chen$^{\dagger}$, Yanwei~Fu$^{\dagger}$, Yinda Zhang, Yu-Gang
Jiang$^{\star}$, Xiangyang Xue, and Leonid~Sigal \thanks{$^{\star}$: Corresponding Author; $^{\dagger}$: Co-first author}
\thanks{Zitian Chen, Yanwei Fu, Xiangyang Xue and Yu-Gang Jiang are with the Schools of Data Science and Computer Science, Fudan University, Shanghai, China.
	Email: \{chenzt15, yanweifu, xyxue, ygj\}@fudan.edu.cn} 
\thanks{Yu-Gang Jiang is also with Jilian Technology Group (Video++), Shanghai, China.}
\thanks{Yinda Zhang is with Princeton University. Email: yindaz@cs.princeton.edu. } 
\thanks{Leonid Sigal is with the Department of Computer Science, University
	of British Columbia, BC, Canada. Email: lsigal@cs.ubc.ca. }
}
\maketitle
\begin{abstract}
The ability to quickly recognize and learn new visual concepts from
limited samples enables humans to quickly adapt to new tasks and environments.
This ability is enabled by semantic association of novel concepts
with those that have already been learned and stored in memory. Computers
can start to ascertain similar abilities by utilizing a semantic concept
space. A concept space is a high-dimensional semantic space in which
similar abstract concepts appear close and dissimilar ones far apart.
In this paper, we propose a novel approach to one-shot learning that
builds on this core idea. Our approach learns to map a novel sample
instance to a concept, relates that concept to the existing ones in
the concept space and, using these relationships, generates new instances,
by interpolating among the concepts, to help learning. Instead of
synthesizing new image instance, we propose to directly synthesize
instance features by leveraging semantics using a novel auto-encoder
network we call {\em dual TriNet}. The encoder part of the TriNet
learns to map multi-layer visual features from CNN to a semantic vector.
In semantic space, we search for related concepts, which are then
projected back into the image feature spaces by the decoder portion
of the TriNet. Two strategies in the semantic space are explored.
Notably, this seemingly simple strategy results in complex augmented
feature distributions in the image feature space, leading to substantially
better performance. 
\end{abstract}


\section{Introduction}

Recent successes in machine learning, especially deep learning, rely
greatly on the training processes that operate on hundreds, if not
thousands, of labeled training instances for each class. However,
in practice, it might be extremely expensive or even infeasible to
obtain many labelled samples, \textit{e.g.} for rare objects or objects
that may be hard to observe. In contrast, humans can easily learn
to recognize a novel object category after seeing only few training
examples \cite{Thrun96learningto}. Inspired by this ability, few-shot
learning aims to build classifiers from few, or even a single, examples.

The major obstacle of learning good classifiers in a few-shot learning
setting is the lack of training data. Thus a natural recipe for few-shot
learning is to first augment the data in some way. A number of approaches
for data augmentation have been explored. The dominant approach, adopted
by the previous work, is to obtain more images \cite{KrizhevskySH12}
for each category and use them as training data. These additional
augmented training images could be borrowed from unlabeled data \cite{transductiveEmbeddingJournal}
or other relevant categories \cite{yuxiong2016eccv,yuxiong2016nips,zhizhong2016eccv,lim2011nips}
in an unsupervised or semi-supervised fashion. However{, the augmented
data that comes from related classes is often semantically noisy and
can result in the \emph{negative transfer} which leads to reduced
(instead of improved) performance. On the other hand, synthetic images
rendered from virtual examples \cite{Movshovitz2015phd,Park2015cvpr,attias2015cvpr,Dosovitskiy2015cvpr,zhu2015ijcv,opelt2006alphabet}
are semantically correct but require careful domain adaptation to
transfer the knowledge and features to the real image domain. To avoid
the difficulty of generating the synthesized images directly, it is
thus desirable to augment the samples in the feature space itself.
For example, the state-of-the-art deep Convolutional Neural Networks
(CNNs) stack multiple feature layers in a hierarchical structure;
{we hypothesize that feature augmentation can, in this case, be done
in feature spaces produced by CNN layers.

Despite clear conceptual benefits, feature augmentation techniques
have been relatively little explored. The few examples include \cite{zhu2015ijcv,opelt2006alphabet,AGA_2017}.
Notably, \cite{zhu2015ijcv} and \cite{opelt2006alphabet} employed
the feature patches (\emph{e.g.} HOG) of the object parts and combined
them to synthesize new feature representations. Dixit {\em et al.}
\cite{AGA_2017}, for the first time, considered attributes-guided
augmentation to synthesize sample features. Their work, however, utilizes
and relies on a set of pre-defined semantic attributes.

A straightforward approach to augment the image feature representation
is to add random (vector) noise to a representation of each single
training image. However, such simple augmentation procedure may not
substantially inform/improve the decision boundary. Human learning
inspires us to search for related information in the concept space.
Our key idea is to leverage additional semantic knowledge, \emph{e.g.}
encapsulated by the semantic space pre-trained using the linguistic
model such as Google's word2vec \cite{distributedword2vec2013NIPS}.
In such semantic manifold similar concepts tend to have similar semantic
feature representations. The overall space demonstrates semantic continuity,
which makes it ideal for feature augmentation.

To leverage such semantic space, we propose a dual TriNet architecture
($g\left(\mathbf{x}\right)=g_{Dec}\circ g_{Enc}\left(\mathbf{x}\right)$)
to learn the transformation between the {image} features {at}
multiple layers and the semantic {space}. The dual TriNet is {paired}
with the 18-layer residual net (ResNet-18) \cite{he2015deep}; it
has encoder TriNet ($g_{Enc}(\mathbf{x})$) and the decoder TriNet
($g_{Dec}(\mathbf{x})$). Specifically, given one training instance,
we can use the ResNet-18 to extract the features at different layers.
The $g_{Enc}(\mathbf{x})$ efficiently maps these features into the
semantic {space}. In the semantic space, the projected instance
features can be corrupted by adding Gaussian noise, or replaced by
its nearest semantic word vectors. {We assume that} slight changes
of feature values {in the semantic space will allow us to maintain
semantic information while spanning the potential class variability.
The decoder TriNet ($g_{Dec}(\mathbf{x})$) is {then} adapted to
map the perturbed semantic instance features back to {multi-layer
(ResNet-18)} feature space. {It is worth noting that Gaussian augmentations/perturbations
in the semantic space ultimately result in highly non-Gaussian augmentations
in the original feature space. This is the core benefit of the semantic
space augmentation. Using three classical supervised classifiers,
we show that the augmented features can boost the performance in few-shot
classification.

\noindent \textbf{Contributions}. Our contributions are in several
fold. First, we propose a simple and yet elegant deep learning architecture:
ResNet-18+dual TriNet with an efficient end-to-end training for few-shot
classification. Second, we illustrate that the proposed dual TriNet
can effectively augment visual features produced by multiple layers
of ResNet-18. Third, and interestingly, we show that we can utilize
semantic spaces of various types, including semantic attribute space,
semantic word vector space, or even subspace defined by the semantic
relationship of classes. Finally, extensive experiments on four datasets
validate the efficacy of the proposed approach in addressing the few-shot
image recognition task.

\section{Related work}

\subsection{Few-Shot Learning}

\noindent Few-shot learning is inspired by human ability to learn
new concepts from very few examples \cite{Jankowski,compositional_1shot}.
Being able to recognize and generalize to new classes with only one,
or few, examples \cite{bart2005cross_gen} is beyond the capabilities
of typical machine learning algorithms, which often rely on hundreds
or thousands of training examples. Broadly speaking there are two
categories of approaches for addressing such challenges:

\vspace{0.05in}
\noindent  \textbf{Direct supervised learning-based approaches,} directly learn
a one-shot classifier via instance-based learning (such as K-nearest
neighbor), non-parametric methods \cite{feifei2003unsup_1s_objcat_learn,feifei2006one_shot,tommasi2009transfercat},
deep generative models \cite{generative_1shot,deep_1shot_recent},
or Bayesian auto-encoders \cite{kingama2014iclr}. Compared with our
work, these methods employ a rich class of generative models to explain
the observed data, rather than directly augmenting instance features
as proposed.

 \vspace{0.05in}

\noindent \textbf{Transfer learning-based approaches,} are explored
via the paradigm of learning to learn \cite{Thrun96learningto} or
meta-learning \cite{JVilalta2002AIR}. Specifically, these approaches
employ the knowledge from auxiliary data to recognize new categories
with few examples by either sharing features \cite{bart2005cross_gen,hertz2016icml,Fleuret2005nips,amit2007icml,wolfc2005cvpr,torralba2005pami},
semantic attributes \cite{lampert13AwAPAMI,transferlearningNIPS,rohrbach2010semantic_transfer},
or contextual information \cite{one_shot_TL_contexutal}. Recently,
the ideas of learning metric spaces from source data to support one-shot
learning were quite extensively explored. Examples include matching
networks \cite{matchingnet_1shot} and prototypical networks \cite{prototype_network}.
Generally, these approaches can be roughly categorized as either meta-learning
algorithms (including MAML {\cite{MAML}, Meta-SGD {\cite{meta-sgd},
DEML+Meta-SGD \cite{DEML+Meta-SGD}, META-LEARN LSTM {\cite{Sachin2017},
Meta-Net \cite{MetaNetwork}, R2-D2\cite{closedform}, Reptile\cite{DBLP:journals/corr/abs-1803-02999},
WRN~\cite{predictFromActivation}) and metric-learning algorithms (including{
}Matching Nets {\cite{matchingnet_1shot}, }PROTO-NET {\cite{prototype_network}},
RELATION NET \cite{relation_net},  MACO {\cite{2018arXiv180204376H},
and \textcolor{black}{Cos \& Att. }\textcolor{black}{\cite{dym}}). In 
\cite{zhongwen2016,memorymatching}, they maintained external memory for
continuous learning. MAML \cite{pmlr-v70-finn17a} can learn good
initial neural network weights which can be easily fine-tuned for
unseen categories. The \cite{2017arXiv171104043G} used graph neural
network to perform message-passing inference from support images to
test images. TPN~\cite{TPN} proposed a framework for transductive
inference thus to solve the data-starved problem.\textcolor{black}{{}
Multi-Attention~\cite{multiAttention} utilized semantic information
to generate attention map to help one-shot recognition, whereas we
directly augment samples in the semantic space and then map them back
to the visual space.} With respect to these works, our framework is
orthogonal but potentially useful -- it is useful to augment instance
features of novel classes before applying such methods.

\subsection{Augmenting training instances\label{subsec:One-shot-learning-by}}

The standard augmentation techniques are often directly applied in
the image domain, such as flipping, rotating, adding noise and randomly
cropping images \cite{KrizhevskySH12,returnDevil2014BMVC,visualizing_network}.
Recently, more advanced data augmentation techniques have been studied
to train supervised classifiers. In particular, {augmented} training
data can also be employed to alleviate the problem of instances {scarcity}
and thus avoid overfitting in one-shot/few-shot learning settings.
{P}revious approaches {can be categorized into six classes of methods}:
(1) Learning one-shot models by utilizing the manifold information
of a large amount of unlabelled data in a semi-supervised or transductive
{setting} \cite{transductiveEmbeddingJournal}; (2) Adaptively learning
the one-shot classifiers from off-shelf trained models \cite{yuxiong2016eccv,yuxiong2016nips,zhizhong2016eccv};
(3) Borrowing examples from relevant categories \cite{lim2011nips,Delta-encoder}
or semantic vocabularies \cite{ssvoc_2016_CVPR,deep_0shot} to augment
the training set; (4) Synthesizing additional labelled training instances
by rendering virtual examples \cite{Movshovitz2015phd,Park2015cvpr,attias2015cvpr,Dosovitskiy2015cvpr,renderCNN}
or composing synthesized representations \cite{zhu2015ijcv,opelt2006alphabet,gait_augmentation,mocap_guide,detector_3d,2017ICCVaug}
or distorting existing training examples \cite{KrizhevskySH12}; (5)
Generating new examples {using} Generative Adversarial Networks
(GANs) \cite{gan_cycle,zhu2015eccv,gan2014,reed2016generative,dcgan,LSgan,ishan2017iclr,xun2017cvpr,imaginaryData};
\textcolor{black}{(6) Attribute-guided augmentation (AGA) and Feature
Space Transfer~\cite{AGA_2017,FSTransfer} to synthesize sample{s}
at desired values, poses or strength.}

{Despite the breadth of research,} previous methods may suffer from
several problems: (1) semi-supervised algorithms {rely on} the manifold
assumption, which, however, cannot {be} effectively validated {in
practice}. (2) {transfer learning} may suffer from the \emph{negative
transfer }when the off-shell models or relevant categories are very
different from one-shot classes; (3) rendering, composing or distorting
{existing} training examples {may require domain expertise;} (4)
GAN-based approaches mostly focuse on learning good generators to
synthesize ``realistic'' images to ``cheat'' the discriminators.
{S}ynthesized images may not necessarily preserve the discriminative
information. This is in contrast to our network structure, where the
discriminative instances are directly synthesized in visual feature
domain. The AGA~\cite{AGA_2017} mainly employed the attributes of
3D depth or pose information for augmentation; in contrast, our methods
can additionally utilize semantic {information} to augment data.
Additionally, the proposed dual TriNet networks can effectively augment
multi-layer features.

\subsection{Embedding Network structures}

Learning {of} visual-semantic embedding{s} has been explored in
various ways, including {with} neural networks, \emph{e.g.}, Siamese
network \cite{Bromley1993ijcai,siamese_1shot}, discriminative methods
(\emph{e.g.}, Support Vector Regressors (SVR) \cite{lampert13AwAPAMI,farhadi2009attrib_describe,Kienzle2006icml}),
metric learning methods \cite{matchingnet_1shot,quattoni2008sparse_transfer,fink2005nips},
or kernel embedding methods \cite{hertz2016icml,wolf2009iccv}. One
of the most common embedding approaches is to project visual features
and semantic entities into a common {\em new} space. However, when
dealing with the feature space of different layers in CNNs, previous
methods have to learn an individual visual semantic embedding for
each layer. In contrast, the proposed Dual TriNet can effectively
learn a single visual-semantic embedding for multi-layer feature spaces.

Ladder Networks \cite{ladderNet} utilize the lateral connections
as auto-encoders for semi-supervised learning tasks. In \cite{deeplyfused}, the authours
fused different intermediate layers of different networks to improve
the image classification performance. Deep Layer Aggregation \cite{layer_aggregation}
aggregated the layers and blocks across a network to better fuse the
information across layers. Rather than learn a specific aggregation
node to merge different layers, our dual TriNet directly transforms,
rescales and concatenates the features of different layers in an encoder-decoder
structure.

\section{Dual TriNet Network for Semantic Data Augmentation}

\begin{figure*}
\centering{}\includegraphics[scale=0.45]{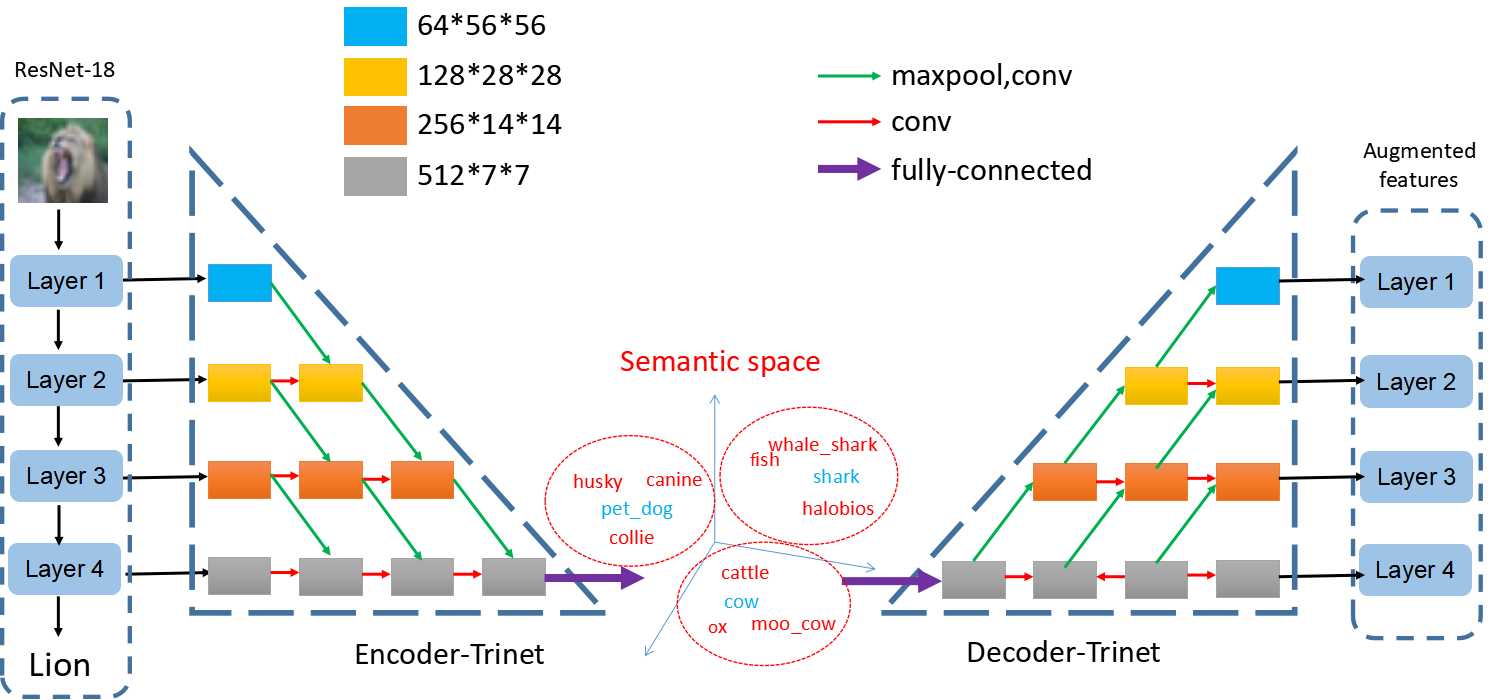}\caption{\label{fig:Overview-of-our}\textbf{Overview of our framework.} We
extract image features by ResNet-18 and augment features by dual TriNet.
Encoder TriNet projects features to the semantic space. After augmenting
data in semantic space, we use the decoder TriNet to obtain the corresponding
augmented features. Both real and augmented data are used to train
the classification model. Note that: (1) the small green arrow indicates
the max pooling with $2\times2$, and following by a ``conv'' layer
which is the sequence Conv-BN-ReLU.}
\end{figure*}

\subsection{Problem setup\label{subsec:Problem-setup}}

In one-shot learning, we are given the base categories $C_{base}$,
and novel categories $C_{novel}(C_{base}\bigcap C_{novel}=\emptyset)$
with the total class label set $\mathcal{C}=\mathcal{C}_{base}\cup\mathcal{C}_{novel}$.
The base categories $C_{base}$ have sufficient labeled image data
and we assume the base dataset $D_{base}=\left\{ \mathbf{I}_{i}^{base},z_{i}^{base},\mathbf{u}_{z_{i}}^{base}\right\} _{i=1}^{N_{base}}$
of $N_{base}$ samples. $\mathbf{I}_{i}^{base}$ indicates the raw
image $i$; $z_{i}^{base}\in\mathcal{C}_{base}$ is a class label
from the base class set; $\mathbf{u}_{z_{i}}^{base}$ is the semantic
vector of the instance $i$ in terms of its class label. The semantic
vector $\mathbf{u}_{z_{i}}^{base}$ can be either semantic attribute
\cite{lampert13AwAPAMI}, semantic word vector \cite{distributedword2vec2013NIPS}
or any representation obtained in the subspace constructed or learned
from semantic relationship of classes.

For novel categories $C_{novel}$, we consider the another dataset
$D_{novel}=\left\{ \mathbf{I}_{i}^{novel},z_{i}^{novel},\mathbf{u}_{z_{i}}^{novel}\right\} $
and each class $z_{i}^{novel}\in\mathcal{C}_{novel}$ . For the novel
dataset, we have a support set and test set. Support set $D_{support}=\left\{ \mathbf{I}_{i}^{support},z_{i}^{support},\mathbf{u}_{z_{i}}^{support}\right\} $
($D_{support}\in D_{novel}$) is composed of a small number of training
instances of each novel class. The test set $D_{test}=\left\{ \mathbf{I}_{i}^{test},z_{i}^{test},\mathbf{u}_{z_{i}}^{test}\right\} $
($D_{test}\in D_{novel},D_{support}\bigcap D_{test}=\emptyset$) is
not available for training, but is used for testing. In general, we
only train on $D_{base}$ and $D_{support}$ ,which contain adequate
instances of base classes and a small number of instances of novel
classes respectively. Then we evaluate our model on $D_{test}$, which
only consists of novel classes. We target learning a model that can
generalize well to the novel categories, 
by using only a small support set $D_{support}$.

\subsection{Overview\label{subsec:Overview}}

\noindent \textbf{Objective.} We seek to directly augment the features
of the training instances of each target class. Given one training
instance $\mathbf{I}_{i}^{support}$ from the novel classes, the feature
extractor network can output the instance feature $\left\{ f_{l}\left(\mathbf{I}_{i}^{support}\right)\right\} $
($l=1,\cdots,L$); and the augmentation network $g\left(\mathbf{x}\right)$
can generate a set of synthesized features $g\left(\left\{ f_{l}\left(\mathbf{I}_{i}^{support}\right)\right\} \right)$.
Such synthesized features are used as additional training instances
for one-shot learning. As illustrated in Fig. \ref{fig:Overview-of-our},
we use the ResNet-18 \cite{he2015deep} and propose a Dual TriNet
network as the feature extractor network and the augmentation network
respectively. The whole architecture is trained in an end-to-end manner
by combining the loss functions of both networks, 
\begin{equation}
\left\{ \Omega,\Theta\right\} =\underset{\Omega,\Theta}{\mathrm{argmin}}J_{1}\left(\Omega\right)+\lambda\cdot J_{2}\left(\Theta\right)\label{eq:jointly_loss}
\end{equation}
where $J_{1}\left(\Omega\right)$ and $J_{2}\left(\Theta\right)$
are the loss functions for ResNet-18 \cite{he2015deep} and dual TriNet
network respectively; $\Omega$ and $\Theta$ represent corresponding
parameters. The cross entropy loss is used for $J_{1}\left(\Omega\right)$
as in~\cite{he2015deep}. Eq.~(\ref{eq:jointly_loss}) is optimized
using base dataset $D_{base}$.

 \vspace{0.05in}

\noindent \textbf{Feature extractor network.} We train 
ResNet-18~\cite{he2015deep} to convert the raw images into feature
vectors. ResNet-18 has 4 sequential residual layers, \emph{i.e.},
layer1, layer2, layer3 and layer4 as illustrated in Fig.~\ref{fig:Overview-of-our}.
Each residual layer outputs a corresponding feature map $f_{l}\left(\mathbf{I}_{i}\right),\quad l=1,\dots4$.
If we consider each feature map a different image representation,
ResNet-18 actually learns a Multi-level Image Feature (M-IF) encoding.
Generally, different layer features may be used for various one-shot
learning tasks. For example, as in \cite{KrizhevskySH12}, the features
of fully connected layers can be used for one-shot image classification;
and the output features of fully convolutional layers may be preferred
for one-shot image segmentation tasks \cite{fully_conv_seg,one_shot_seg17BMVC,one-shot-video17cvpr}.
By combining features from multiple levels, our method can be applied
to a variety of different visual tasks.

\vspace{0.05in}

\noindent \textbf{Augmentation network.} We propose an encoder-decoder
architecture -- dual TriNet ($g\left(\mathbf{x}\right)=g_{Dec}\circ g_{Enc}\left(\mathbf{x}\right)$).
As illustrated in Fig.~\ref{fig:Overview-of-our}, our dual TriNet
can be divided into encoder-TriNet $g_{Enc}\left(\mathbf{x}\right)$
and decoder-TriNet sub-network $g_{Dec}\left(\mathbf{x}\right)$.
The encoder-TriNet maps visual feature space to a semantic space.
This is where augmentation takes place. The decoder-TriNet projects
the augmented semantic space representation back to the feature space.
Since ResNet-18 has four layers, the visual feature spaces produced
by different layers can use the same encoder-decoder TriNet for data
augmentation.

\subsection{Dual TriNet Network}

The dual TriNet is paired with ResNet-18. Feature representations
obtained from different layers of such a deep CNN architecture, are
hierarchical, going from \emph{general} (bottom layers) to more \emph{specific}
(top layers) \cite{transferable_deep_feat}. For instance, the features
produced by the first few layers are similar to Gabor filters \cite{visualizing_network}
and thus agnostic to the tasks; in contrast, the high-level layers
are specific to a particular task, \emph{e.g}., image classification.
The feature representations produced by layers of ResNet-18 have different
levels of abstract semantic information. Thus a natural question is
whether we can augment features at different layers? Directly learning
an encoder-decoder for each layer will not fully exploit the relationship
of different layers, and thus may not effectively learn the mapping
between feature spaces and the semantic space. To this end, we propose
the dual TriNet network.

Dual TriNet learns the mapping between the Multi-level Image Feature
(M-IF) encoding and the Semantic space. The semantic space can be
either semantic attribute space, or semantic word vector space introduced
in Sec. \ref{subsec:Problem-setup}. Semantic attributes can be pre-defined
by human experts \cite{AGA_2017}. Semantic word vector $\mathbf{u}_{z_{i}}^{base}$
is the projection of each vocabulary entity $w_{i}\in\mathcal{W}$,
where vocabulary $\mathcal{W}$ is learned by word2vec \cite{distributedword2vec2013NIPS}
on a large-scale corpus. Furthermore, the subspace $\mathbf{u}_{z_{i}}^{base}$
can be spanned by Singular Value Decomposition (SVD) of the semantic
relationship of classes. Specifically, we can use $\left\{ \mathbf{u}_{z_{i}}^{base};\mathbf{u}_{z_{j}}^{novel}\right\} _{z_{i}\in\mathcal{C}_{base},z_{j}\in\mathcal{C}_{novel}}$
to compute the semantic relationship $\mathbf{M}$ of classes using
cosine similarity. We decompose $\mathbf{M}=\mathbf{U}\mathbf{\Sigma}\mathbf{V}$
by SVD algorithm. The $\mathbf{U}$ is a unitary matrix and defines
a new semantic space. Each row of $\mathbf{U}$ is taken as a new
semantic vector of one class.

Encoder TriNet is composed of four layers corresponding to each layer
of ResNet-18. It aims to learn the function $\hat{\mathbf{u}}_{z_{i}}=g_{Enc}\left(\left\{ f_{l}\left(\mathbf{I}_{i}\right)\right\} \right)$
to map all layer features $\left\{ f_{l}\left(\mathbf{I}_{i}\right)\right\} $
of instance $i$ as close to the semantic vector $\mathbf{u}_{z_{i}}$
of instance $i$ as possible. The structure of subnetwork is inspired
by the tower of Hanoi as shown in Fig.~\ref{fig:Overview-of-our}.
Such a structure can efficiently exploit the differences and complementarity
of information encoded in multiple layers. The encoder TriNet is trained
to match the four layers of ResNet-18 by merging and combining the
outputs of different layers. The decoder TriNet has inverse architecture
to project the features $\hat{\mathbf{u}}_{z_{i}}$ from semantic
space to the feature space $\hat{f_{l}}\left(\mathbf{I}_{i}\right)=g_{Dec}\left(g_{Enc}\left(\left\{ f_{l}\left(\mathbf{I}_{i}\right)\right\} \right)\right)$.
We learn TriNet by optimizing the following loss: 
\begin{equation}
\;J_{2}\left(\Theta\right)=\mathbb{E}\left[\sum_{l=1}^{4}\left(f_{l}\left(\mathbf{I}_{i}\right)-\hat{f_{l}}\left(\mathbf{I}_{i}\right)\right)^{2}+\left(\hat{\mathbf{u}}_{z_{i}}-\mathbf{u}_{z_{i}}\right)^{2}\right]+\lambda P\left(\Theta\right)\label{eq:auto_encoder}
\end{equation}

\noindent \textcolor{black}{where $\mathbf{I}_{i}\in D_{base}$ and
$\Theta$ indicates the parameter set of dual TriNet network }and
$P\left(\cdot\right)$ is the $L_{2}-$regularization term. The dual
TriNet is trained on $D_{base}$ and used to synthesize instances
in the form of $l$-th layer feature perturbations with respect to
a given training instance from $D_{support}$.

\subsection{Feature Augmentation by Dual TriNet \label{subsec:Data-Augmentation-by}}

With the learned dual TriNet, we have two ways to augment the features
of training instances. Note that the augmentation method is only used
to extend $D_{support}$.

\vspace{0.05in}

\noindent \textbf{Semantic Gaussian (SG).} A natural way to augment
features is by sampling instances from a Gaussian distribution. Specifically,
for the feature set $\left\{ f_{l}\left(\mathbf{I}_{i}^{support}\right)\right\} $($l=1,\cdots,L$)
extracted by ResNet-18, the encoder TriNet can project the $\left\{ f_{l}\left(\mathbf{I}_{i}^{support}\right)\right\} $
into the semantic space, $g_{Enc}\left(\left\{ f_{l}\left(I_{i}^{support}\right)\right\} \right)$.
In such a space, we assume that vectors 
can be corrupted by a random Gaussian noise without changing a semantic
label. 
This can be used to augment the data. Specifically, we sample the
$k-th$ semantic vector $\mathbf{v}_{i}^{G_{k}}$ from $I_{i}^{support}$
using semantic Gaussian as follows, 
\begin{equation}
\mathbf{v}_{i}^{G_{k}}\sim\mathcal{N}\left(g_{Enc}\left(\left\{ f_{l}\left(\mathbf{I}_{i}^{support}\right)\right\} \right),\sigma\mathbf{E}\right)\label{eq:semantic_gaussian}
\end{equation}
where $\sigma\in\mathbb{R}$ is the variance of each dimension and
$\mathbf{E}$ is the identity matrix; $\sigma$ controls the standard
deviation of the noise being added. To make the augmented semantic
vector $\mathbf{v}_{i}^{G_{k}}$ still be representative of the class
of $z_{i}^{support}$, we empirically set $\sigma$ to $15\%$ of
the distance between $u_{z_{i}}^{support}$ and its nearest other
class instance $\mathbf{u}_{z_{j}}^{support}$ ($z_{i}^{support}\neq z_{j}^{support}$)
as this gives the best performance. The decoder TriNet generates the
virtual synthesized sample $g_{Dec}\left(\mathbf{v}_{i}^{G_{k}}\right)$
which shares the same class label $z_{i}^{support}$ with the original
instance. By slightly corrupting the values of some dimensions of
semantic vectors, we expect the sampled vectors $\mathbf{\mathbf{v}}_{i}^{G_{k}}$
to still have the same semantic meaning. 

\vspace{0.05in}

\noindent \textbf{Semantic Neighborhood (SN).} Inspired by the recent
work on vocabulary-informed learning \cite{ssvoc_2016_CVPR}, the
large amount of vocabulary in the semantic word vector space (\textit{e.g.},
word2vec \cite{distributedword2vec2013NIPS}) can also be used for
augmentation. The distribution of such vocabulary reflects the general
semantic relationships in the linguistic corpora. For example, in
word vector space, the vector of ``truck'' is closer to the vector
of ``car'' than to the vector of ``dog''. Given the features $\left\{ f_{l}\left(\mathbf{I}_{i}^{support}\right)\right\} $
of training instance $i$, the $k$-th augmented data $\mathbf{v}_{i}^{N_{k}}$
can be sampled from the neighborhood of\emph{ }$g_{Enc}\left(\left\{ f_{l}\left(\mathbf{I}_{i}^{support}\right)\right\} \right)$,
\emph{i.e.}, 
\begin{equation}
\mathbf{v}_{i}^{N_{k}}\in Neigh\left(g_{Enc}\left(\left\{ f_{l}\left(\mathbf{I}_{i}^{support}\right)\right\} \right)\right)\label{eq:semantic_neighbourhood}
\end{equation}
$Neigh\left(g_{Enc}\left(\left\{ f_{l}\left(\mathbf{I}_{i}^{support}\right)\right\} \right)\right)\subseteq\mathcal{W}$
indicates the nearest neighborhood vocabulary set of $g_{Enc}\left(\left\{ f_{l}\left(\mathbf{I}_{i}^{t}\right)\right\} \right)$
and $\mathcal{W}$ indicate vocabulary set learned by word2vec \cite{distributedword2vec2013NIPS}
on a large-scale corpus. These neighbors correspond to the most semantically
similar examples to our training instance. The features of synthesized
samples can again be decoded by $g_{Dec}\left(\mathbf{v}_{i}^{N_{k}}\right)$.

There are several points we want to highlight. (1) For one training
instance $\mathbf{I}_{i}^{support}$, we use as the Gaussian mean
in Eq (\ref{eq:semantic_gaussian}) or neighborhood center in Eq (\ref{eq:semantic_neighbourhood}),
the $g_{Enc}\left(\left\{ f_{l}\left(\mathbf{I}_{i}^{support}\right)\right\} \right)$
rather than its ground-truth word vector $\mathbf{u}_{z_{i}}^{support}$.
This is due to the fact that $\mathbf{u}_{z_{i}}^{support}$ only
represents the semantic center of class $z_{i}^{support}$, not the
center for the instance $i$. Experimentally, on \emph{mini}ImageNet
dataset, augmenting features using $\mathbf{u}_{z_{i}}^{support}$,
rather than $g_{Enc}\left(\left\{ f_{l}\left(\mathbf{I}_{i}^{t}\right)\right\} \right)$,
leads to $3\sim5\%$ performance drop (on average) in 1-shot/5-shot
classification. (2) Semantic space Gaussian noise added in Eq~(\ref{eq:semantic_gaussian})
or semantic neighborhood used in Eq (\ref{eq:semantic_neighbourhood})
result in the synthesized training features that are highly nonlinear
(non-Gaussian) for each class. This is the result of non-linear decoding
provided by TriNet $g_{Dec}\left(\mathbf{x}\right)$ and ResNet-18
($\left\{ f_{l}\left(\mathbf{I}_{i}^{t}\right)\right\} $). (3) Directly
adding Gaussian noise to $\left\{ f_{l}\left(\mathbf{I}_{i}^{t}\right)\right\} $
is another naive way to augment features. However, in \emph{mini}ImageNet
dataset, such a strategy does not give any significant improvement
in one-shot classification.

\subsection{One-shot Classification\label{subsec:One-shot-Classification-and}}

Having trained feature extractor network and dual TriNet on base dataset
$D_{base}$, we now discuss conducting one-shot classification on
the target dataset $D_{novel}$. For the instance $i$ in $D_{novel}$
we can extract the M-IF representation $f_{l}\left(\mathbf{I}_{i}^{novel}\right)$
($l=1,2,...,L$) using the feature extractor network. We then use
the encoder part of TriNet to map all layer features $\left\{ f_{l}\left(\mathbf{I}_{i}\right)\right\} $
of instance $i$ to semantic vector $g_{Enc}\left(\left\{ f_{l}\left(\mathbf{I}_{i}^{support}\right)\right\} \right)$.

Our framework can augment the instance, producing multiple synthetic
instances in addition to the original one $\left\{ \mathbf{v}_{i}^{G_{k}}\right\} \cup\left\{ \mathbf{v}_{i}^{N_{k}}\right\} $
using semantic Gaussian and/or semantic neighborhood approaches discussed.
For each new semantic vector $\mathbf{v}_{i}^{k}\in\left\{ \mathbf{v}_{i}^{G_{k}}\right\} \cup\left\{ \mathbf{v}_{i}^{N_{k}}\right\} $,
we use decoder TriNet to map them from semantic space to all layer
features $\left\{ x_{l}^{augment_{i}}\right\} =g_{Dec}(v_{i}^{k})$
($l=1,2,...,L$). The features that are not at the final $L$-th layer
are feed through from $l+1$-th layer to $L$-th layer of feature
extractor network to obtain $\left\{ \hat{x}_{l}^{augment}\right\} $.
Technically, one new semantic vector $\mathbf{v}_{i}^{k}$ can generate
$L$ instances: one from each of the $L$ augmented layers. Consistent
with previous work~\cite{KrizhevskySH12,he2015deep}, the features
produced by the final layer are utilized for one-shot classification
tasks. Hence the newly synthesized $L$-th layer features $\left\{ \hat{x}_{l}^{augment_{i}}\right\} $
obtained from the instance $i$ and original $L$-th layer feature
$f_{L}\left(\mathbf{I}_{i}^{support}\right)$ are used to train the
one-shot classifier $g_{one-shot}(x)$ in a supervised manner. Note
that all augmented feature vectors obtained from instance $i$ are
assumed to have the same class label as the original instance $i$.

In this work, we show that the augmented features can benefit various
supervised classifiers. To this end three classical classifiers, \emph{i.e.},
the K-nearest neighbors (KNN), Support Vector Machine (SVM) and Logistic
Regression (LR), are utilized as one-shot classifier $g_{one-shot}(x)$.
In particular, we use $g_{one-shot}(x)$ to classify the $L$-th layer
feature $f_{l}\left(\mathbf{I}_{i}^{test}\right)$ of test sample
$\mathbf{I}_{i}^{test}$ at the test time.

\section{Experiments}

\subsection{Datasets}

We conduct experiments on four datasets. Note that (1) on all datasets,
ResNet-18 is only trained on the training set (equivalent to base
dataset) in the specified splits of previous works. (2) The same networks
and parameter settings (including the size of input images) are used
for all the datasets; hence all images are resized to $224\times224$.

 \vspace{0.05in}

\noindent \textbf{\emph{mini}}\textbf{ImageNet.} Originally proposed
in \cite{matchingnet_1shot}, this dataset has 60,000 images from
100 classes; each class has around 600 examples. To make our results
comparable to previous works, we use the splits in \cite{Sachin2017}
by utilizing 64, 16 and 20 classes for training, validation and testing
respectively.

 \vspace{0.05in}

\noindent \textbf{Cifar-100.} Cifar-100 contains 60,000 images from
100 fine-grained and 20 coarse-level categories \cite{Krizhevsky2009LearningML}.
We use the same data split as in \cite{2018arXiv180203596Z} to enable
the comparison with previous methods. In particular, 64, 16 and 20
classes are used for training, validation and testing respectively.

\vspace{0.05in}

\noindent \textbf{Caltech-UCSD Birds 200-2011 (CUB-200).} CUB-200
is a fine-grained dataset consisting of a total of 11,788 images from
200 categories of birds \cite{WahCUB_200_2011}. As the split in \cite{2018arXiv180204376H},
we use 100, 50 and 50 classes for training, validation and testing.
This dataset also provides 312 dimensional semantic attribute vectors
on a per-class level.

\vspace{0.05in}

\noindent \textbf{Caltech-256.} Caltech-256 has 30,607 images from
256 classes \cite{griffin2007caltech}. As in \cite{2018arXiv180203596Z},
we split the dataset into 150, 56 and 50 classes for training, validation
and testing respectively.

\begin{table*}
\centering{}%
\begin{tabular}{c|c|c|c|c}
\hline 
\multirow{2}{*}{Methods} & \multicolumn{2}{l|}{\emph{mini}ImageNet{} ($\%$)} & \multicolumn{2}{l}{{}CUB-200($\%$)}\tabularnewline
\cline{2-5} 
 & {}1-shot  & {}5-shot  & {}1-shot  & {}5-shot \tabularnewline
\hline 
\hline 
{}META-LEARN LSTM \cite{Sachin2017}  & {}43.44\textpm 0.77  & {}60.60\textpm 0.71  & {}40.43  & {}49.65 \tabularnewline
\hline 
{}MAML \cite{MAML}  & {}48.70\textpm 1.84  & {}63.11\textpm 0.92  & {}38.43  & {}59.15 \tabularnewline
\hline 
{}Meta-Net \cite{MetaNetwork}  & {}49.21\textpm 0.96  & -  & {}-  & {}- \tabularnewline
\hline 
\textcolor{black}{Reptile\cite{DBLP:journals/corr/abs-1803-02999}} & \textcolor{black}{49.97} & \textcolor{black}{65.99} & \textcolor{black}{-} & \textcolor{black}{-}\tabularnewline
\hline 
MAML{*} \cite{MAML}  & 52.23\textpm 1.24  & 61.24\textpm 0.77  & -  & -\tabularnewline
\hline 
Meta-SGD{*} \cite{meta-sgd}  & 52.31\textpm 1.14  & 64.66\textpm 0.89  & -  & -\tabularnewline
\hline 
{}DEML+Meta-SGD \cite{DEML+Meta-SGD}  & \textbf{{}}58.49{}\textpm 0.91\textbf{{}}  & {}71.28\textpm 0.69  & {}-  & {}- \tabularnewline
\hline 
\hline 
{}MACO \cite{2018arXiv180204376H}  & {}41.09\textpm 0.32  & {}58.32\textpm 0.21  & {}60.76  & {}74.96 \tabularnewline
\hline 
Matching Nets{*} \cite{matchingnet_1shot}  & 47.89\textpm 0.86  & 60.12\textpm 0.68  & -  & -\tabularnewline
\hline 
{}PROTO-NET \cite{prototype_network}  & {}49.42\textpm 0.78  & {}68.20\textpm 0.66  & {}45.27  & {}56.35 \tabularnewline
\hline 
\textcolor{black}{GNN~\cite{2017arXiv171104043G}} & \textcolor{black}{50.33\textpm 0.36} & \textcolor{black}{66.41\textpm 0.63} & \textcolor{black}{-} & \textcolor{black}{-}\tabularnewline
\hline 
\textcolor{black}{R2-D2~\cite{closedform}} & \textcolor{black}{51.5\textpm 0.2 } & \textcolor{black}{68.8\textpm 0.1 } & \textcolor{black}{-} & \textcolor{black}{-}\tabularnewline
\hline 
\textcolor{black}{MM-Net~\cite{memorymatching}} & \textcolor{black}{53.37\textpm 0.48} & \textcolor{black}{66.97\textpm 0.35} & \textcolor{black}{-} & \textcolor{black}{-}\tabularnewline
\hline 
\textcolor{black}{Cos \& Att.~\cite{dym}} & \textcolor{black}{55.45\textpm 0.89} & \textcolor{black}{70.13 \textpm 0.68} & \textcolor{black}{-} & \textcolor{black}{-}\tabularnewline
\hline 
\textcolor{black}{TPN~\cite{TPN}} & \textcolor{black}{55.51} & \textcolor{black}{69.86} & \textcolor{black}{-} & \textcolor{black}{-}\tabularnewline
\hline 
SNAIL \cite{SNAIL}  & 55.71\textpm 0.99  & 68.88\textpm 0.92  & -  & -\tabularnewline
\hline 
{}RELATION NET \cite{relation_net}  & {}57.02\textpm 0.92  & {}71.07\textpm 0.69  & {}-  & {}- \tabularnewline
\hline 
\textcolor{black}{Delta-encoder~\cite{Delta-encoder}} & \textcolor{black}{58.7} & \textcolor{black}{73.6} & \textcolor{black}{-} & \textcolor{black}{-}\tabularnewline
\hline 
\textcolor{black}{WRN~\cite{predictFromActivation}} & \textbf{\textcolor{black}{59.60}}\textcolor{black}{\textpm 0.41 } & \textcolor{black}{73.74\textpm 0.19 } & \textcolor{black}{-} & \textcolor{black}{-}\tabularnewline
\hline 
\hline 
{}ResNet-18  & {}52.73\textpm 1.44  & {}73.31\textpm 0.81  & {}66.54\textpm 0.53  & {}82.38\textpm 0.43\tabularnewline
\hline 
ResNet-18+Gaussian Noise  & 52.14\textpm 1.51  & 71.78\textpm 0.89  & 65.02\textpm 0.60  & 80.79\textpm 0.49\tabularnewline
\hline 
{}Ours: ResNet-18+Dual TriNet  & {}58.12\textpm 1.37  & \textbf{{}76.92}{}\textpm 0.69  & \textbf{{}69.61}{}\textpm 0.46  & \textbf{{}84.10}{}\textpm 0.35\tabularnewline
\hline 
\end{tabular}\caption{\label{tab:miniimagenet}Results on \emph{mini}ImageNet and CUB-200.
The ``\textpm '' indicates $95\%$ confidence intervals over tasks.{*}:
indicates the corresponding baselines that are using ResNet-18. Note
that ``\textpm '' is not reported on CUB-200 in previous works.}
\end{table*}

\subsection{Network structures and Settings}

The same ResNet-18 and dual TriNet are used for all four datasets
and experiments.

\vspace{0.05in}

\noindent \textbf{Parameters.} The dropout rate and learning rate
of the auto-encoder network are set to 0.5 and $1e^{-3}$ respectively
to prevent overfitting. The learning rate is divided by 2 every 10
epochs. The batch size is set to 64. The network is trained using
Adam optimizer and usually converges in 100 epochs. To prevent randomness
due to the small training set size, all experiments are repeated multiple
times. 
\textcolor{black}{The Top-1 accuracies are reported with $95\%$ confidence
interval and are averaged over multiple test episodes, the same as
previous work \cite{Sachin2017}. }

\vspace{0.05in}

\noindent \textbf{Settings. }We use the 100-dimensional semantic word
vectors extracted from the vocabulary dictionary released by \cite{ssvoc_2016_CVPR}.
The class name is projected into the semantic space as a vector $\mathbf{u}_{z_{i}}^{base}$
or $\mathbf{u}_{z_{i}}^{novel}$. The semantic attribute space is
pre-defined by experts \cite{lampert13AwAPAMI,WahCUB_200_2011}. In
all experiments, given one training instance 
the dual TriNet will generate 4 augmented instances in the semantic
space. Thus we have 4 synthesized instances of each layer which results
in 
16 synthesized instances in the form of $4-th$ layer features. So
one training instance becomes 17 training instances at the end.

\subsection{Competitors and Classification models}

\noindent \textbf{Competitors}. The previous methods we compare to
are run using the same source/target and training/testing splits as
used by our method. We compare to Matching Nets {\cite{matchingnet_1shot}},
MAML {\cite{MAML}}, Meta-SGD {\cite{meta-sgd},} DEML+Meta-SGD
{\cite{DEML+Meta-SGD}}, PROTO-NET {\cite{prototype_network}},
RELATION NET {\cite{relation_net}}, META-LEARN LSTM {\cite{Sachin2017}},
Meta-Net {\cite{MetaNetwork}}, SNAIL \cite{SNAIL} , MACO \cite{2018arXiv180204376H},
\textcolor{black}{GNN \cite{2017arXiv171104043G}, MM-Net\cite{memorymatching},
Reptile \cite{DBLP:journals/corr/abs-1803-02999}, TPN \cite{TPN},
WRN \cite{predictFromActivation}, Cos \& Att. \cite{dym}, Delta-encoder \cite{Delta-encoder}
and R2-D2\cite{closedform}. To make} a fair comparison, we implement
some of the methods and use ResNet-18 as a commmon backbone architecture.

\vspace{0.05in}

\noindent \textbf{Classification model.} KNN, SVM, and LR are used
as the classification models to validate the effectiveness of our
augmentation technique. \textcolor{black}{The hyperparameters of classification
models are selected using cross-validation on a validation set.}

\subsection{Experimental results on \emph{mini}ImageNet and {CUB-200}}

\begin{table*}
\begin{centering}
\begin{tabular}{c|c|c|c|c|c|c|c|c|c|c|c|c|c|c|c}
\hline 
 & \multicolumn{5}{c|}{Semantic Neighbourhood} & \multicolumn{5}{c|}{Semantic Gaussian} & \multicolumn{5}{c}{Attribute Gaussian}\tabularnewline
\hline 
\hline 
 & 0  & 2  & 4  & 10  & 50  & 0  & 2  & 4  & 10  & 50  & 0  & 2  & 4  & 10  & 50\tabularnewline
\hline 
L1  & 66.5  & 67.1  & 67.2  & \textbf{67.3}  & 67.2  & 66.5  & 67.1  & \textbf{67.2}  & 67.1  & \textbf{67.2}  & 66.5  & \textbf{67.1}  & \textbf{67.1}  & \textbf{67.1}  & \textbf{67.1}\tabularnewline
\hline 
L2  & 66.5  & 67.0  & \textbf{67.1}  & \textbf{67.1}  & 67.0  & 66.5  & \textbf{67.1}  & \textbf{67.1}  & \textbf{67.1}  & 67.0  & 66.5  & 67.1  & \textbf{67.3}  & \textbf{67.3}  & 67.2\tabularnewline
\hline 
L3  & 66.5  & 67.1  & \textbf{67.3}  & \textbf{67.3 } & \textbf{67.3}  & 66.5  & 67.1  & \textbf{67.3}  & \textbf{67.3}  & 67.2  & 66.5  & 67.0  & 67.4  & \textbf{67.5}  & \textbf{67.5}\tabularnewline
\hline 
L4  & 66.5  & 67.4  & \textbf{67.5}  & 67.4  & 67.4  & 66.5  & 67.1  & \textbf{67.3}  & 67.2  & \textbf{67.3}  & 66.5  & 67.1  & \textbf{67.6}  & \textbf{67.6}  & 67.5\tabularnewline
\hline 
M-L  & 66.5  & 68.0  & \textbf{68.1}  & \textbf{68.1}  & \textbf{68.1}  & 66.5  & 67.9  & \textbf{68.0}  & \textbf{68.0}  & \textbf{68.0}  & 66.5  & 68.2  & 68.3  & \textbf{68.4}  & 68.3\tabularnewline
\hline 
\end{tabular}
\par\end{centering}
\caption{\label{tab:augnum}Ablation study of the number of augmented samples
in semantic space on CUB. We report 5-way 1-shot accuracy. L1, L2,
L3, and L4 indicate that we only use the augmented features of Layer
1, Layer 2, Layer 3 and Layer 4 respectively. M-L indicates that we
use all augmented features from four layers; }
\end{table*}

\noindent \textbf{Settings.}{ For }\emph{mini}{ImageNet dataset
we only have a semantic word space. Give one training instance, we
can generate 16 augmented instances for Semantic Gaussian (SG) and
Semantic Neighborhood (SN) each. On CUB-200 dataset, we use both the
semantic word vector and semantic attribute spaces. Hence for one
training instance, we generate 16 augmented instances for SG and SN
each in semantic word vector space; and additionally, we generate
16 virtual instances (in all four layers) for Semantic Gaussian (SG)
in semantic attribute space, which we denote Attribute Gaussian (AG).}

 \vspace{0.05in}
\noindent \textbf{Variants of the number of augmented samples. }Varying the
number of augmented samples does not significantly affect our performance.
To show this, we provide 1-shot accuracy on CUB dataset with the different
numbers of augmented samples (Table~\ref{tab:augnum}). As shown,
the improvements from increasing the number of augmented samples saturate
at a certain point.

\begin{figure*}
\begin{centering}
\includegraphics[scale=0.5]{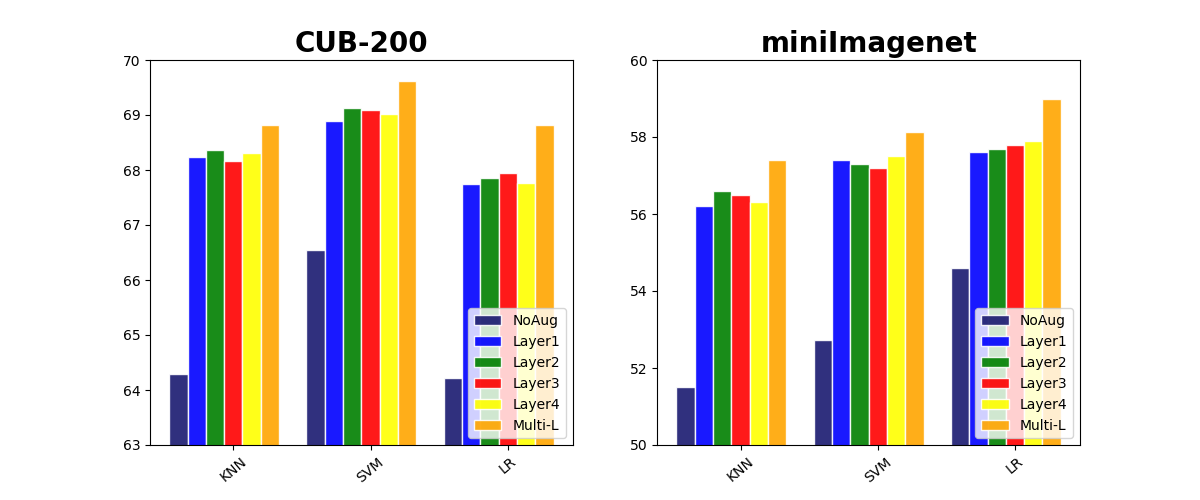} 
\par\end{centering}
\caption{\label{fig:aug-layer} One-shot Results of feature augmentation by
different layers/classifiers on CUB-200 and \emph{mini}ImageNet. ``NoAug'',
``Layer1'', ``Layer2'', ``Layer3'', ``Layer4'' indicate the
one-shot learning results without any augmentation, with the feature
augmentation by using layer 1, layer 2, layer 3, layer 4 of ResNet-18.
``Multi-L'' denotes the performance of using all augmented instances
of one-shot learning. The X-axis represents the different supervised
classifiers. }
\end{figure*}

\vspace{0.05in}

\noindent \textbf{Results. }{As shown in Tab. \ref{tab:miniimagenet},
the competitors can be divided into two categories: Meta-learning
algorithms (including MAML, Meta-SGD, {}DEML+Meta-SGD, META-LEARN
LSTM and Meta-Net)} and Metric-learning algorithms (including Matching
Nets, PROTO-NET, RELATION NET ,SNAIL and MACO). We also report the
results of ResNet-18 (without data augmentation). 
The accuracy of our framework (ResNet-18+Dual TriNet) is also reported.
The Dual TriNet synthesizes each layer features of ResNet-18 as described
in Sec. \ref{subsec:Data-Augmentation-by}. ResNet18+Gaussian Noise
is a simple baseline that synthesizes 16 samples of each test example
by adding Gaussian noise to the $4-th$ layer features. We use SVM
classifiers for ResNet-18 , ResNet18+Gaussian Noise and ResNet-18+Dual
TriNet in Tab. \ref{tab:miniimagenet}. In particular, we found that,

\vspace{0.05in}

\noindent \textbf{(1) Our baseline (ResNet-18) nearly beats all the
other baselines.} Greatly benefiting from learning the residuals,
Resnet-18 is a very good feature extractor for one-shot learning tasks.
Previous works \cite{MAML}\cite{meta-sgd}\cite{matchingnet_1shot}
designed their own network architectures with fewer parameters and
used different objective functions. As can be seen from Tab. \ref{tab:miniimagenet},
after replacing their backbone architecture with ResNet-18, they still
behave worse than our baseline (Resnet-18). We argue that this is
because ResNet-18 is more adaptable to classification task and it
can generate more discriminative space using Cross Entropy Loss than
other objective functions used in metric learning. However, this topic
is beyond our discussion. We want to clarify that since our augmentation
method is capable of being combined with arbitrary approaches, we
choose the strongest baseline to the best of our knowledge. This baseline
can be enhanced by our approach, illustrate the universality of our
augmentation.

\vspace{0.05in}

\noindent \textbf{(2) Our framework can achieve the best performance.}
As shown in Tab. \ref{tab:miniimagenet}, the results of our framework,
\emph{i.e.}, ResNet-18+Dual TriNet can achieve the best performance
and we can show a clear improvements over all the other baselines
on both datasets. This validates the effectiveness of our framework
in solving the one-shot learning task. Note, DEML+Meta-SGD~\cite{DEML+Meta-SGD}
uses the ResNet-50 as the baseline model and hence has better one-shot
learning results than our ResNet-18. Nevertheless, with the augmented
data produced by Dual TriNet we can observe a clear improvement over
ResNet-18. 

\begin{figure*}
\begin{centering}
\includegraphics[scale=0.5]{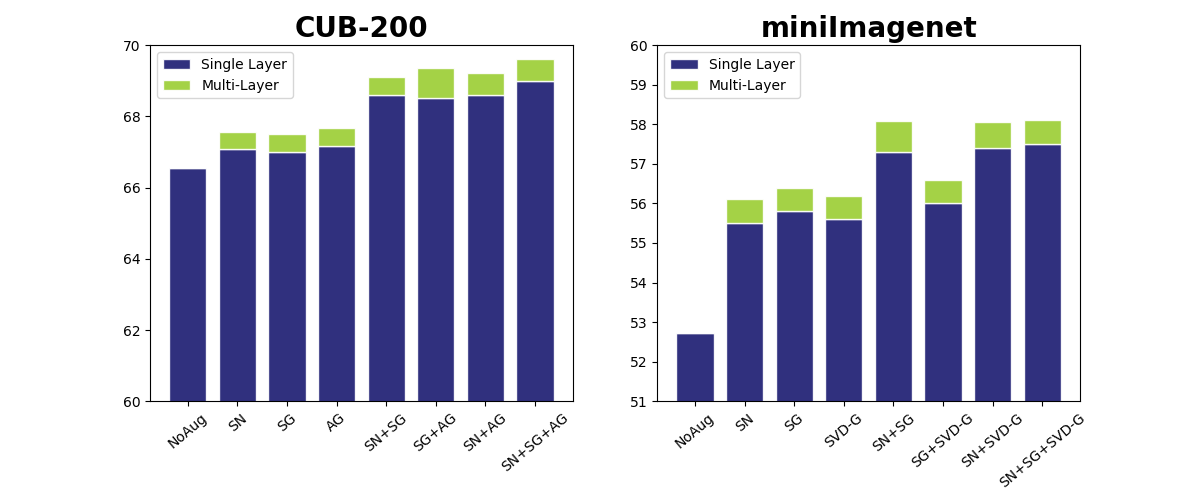} 
\par\end{centering}
\caption{\label{fig:aug-layer-1} One-shot results of feature augmentation
by different types of semantic spaces on CUB-200 and \emph{mini}ImageNet.
``Single Layer'' indicates the best one-shot performance augmented
by using only single layer. ``Multi-layer'' represents the results
of using synthesized instances from all layers.}
\end{figure*}

\vspace{0.05in}

\noindent \textbf{(3) Our framework can effectively augment multiple
layer features. }We analyze the effectiveness of augmented features
in each layer as shown in Fig. \ref{fig:aug-layer}. On CUB-200 and
\emph{mini}ImageNet, we report the results in 1-shot learning cases.
We have several conclusions: (1) Only using the augmented features
from one single layer (\emph{e.g.}, Layer1 -- Layer 4 in Fig. \ref{fig:aug-layer})
can also help improve the performance of one-shot learning results.
This validates the effectiveness of our dual TriNet of synthesizing
features of different layers in a single framework. (2) The results
of using synthesized instances from all layers (Multi-L) are even
better than those of individual layers. This indicates that the augmented
features at different layers are intrinsically complementary to each
other.

\vspace{0.05in}

\noindent \textbf{(4) Augmented features can boost the performance
of different supervised classifiers. }Our augmented features are not
designed for any one supervised classifier. To show this point and
as illustrated in Fig. \ref{fig:aug-layer}, three classical supervised
classifiers (\emph{i.e.}, KNN, SVM and LR) are tested along the X-axis
of Fig. \ref{fig:aug-layer}. Results show that our augmented features
can boost the performance of three supervised classifiers on one-shot
classification cases. This further validates the effectiveness of
our augmentation framework.

\vspace{0.05in}

\noindent \textbf{(5) The augmented features by SG, SN and AG can
also improve few-shot learning results.} We compare different types
of feature augmentation methods of various semantic spaces in Fig.
\ref{fig:aug-layer-1}. Specifically, we compare the SG and SN in
semantic word vector space; and AG in semantic attribute space. On
CUB-200 dataset, the augmented results by SG, SN and AG are better
than those without augmentation. The accuracy of combining the synthesized
instance features generated by any two of SG, SN, and AG can be further
improved over those of SG, SN or AG only. This means that the augmented
feature instances of SG, SN and AG are complementary to each other.
Finally, we observe that by combining augmented instances from all
methods (SG, SN and AG), the accuracy of one-shot learning is the
highest one.

\begin{table*}
\begin{centering}
\begin{tabular}{c|c|c|c|c}
\hline 
\multirow{2}{*}{Methods } & \multicolumn{2}{l|}{{}Caltech-256 ($\%$)} & \multicolumn{2}{l}{{}CIFAR-100 ($\%$)}\tabularnewline
\cline{2-5} 
 & {}1-shot  & {}5-shot  & {}1-shot  & {}5-shot \tabularnewline
\hline 
\hline 
{}MAML \cite{MAML}  & {}45.59\textpm 0.77  & {}54.61\textpm 0.73  & {}49.28\textpm 0.90  & {}58.30\textpm 0.80 \tabularnewline
\hline 
{}Meta-SGD \cite{meta-sgd}  & {}48.65\textpm 0.82  & {}64.74\textpm 0.75  & {}53.83\textpm 0.89  & {}70.40\textpm 0.74 \tabularnewline
\hline 
{}DEML+Meta-SGD \cite{DEML+Meta-SGD}  & {}62.25\textpm 1.00  & {}79.52\textpm 0.63  & {}61.62\textpm 1.01  & {}77.94\textpm 0.74 \tabularnewline
\hline 
\hline 
{}Matching Nets \cite{matchingnet_1shot}  & {}48.09\textpm 0.83  & {}57.45\textpm 0.74  & {}50.53\textpm 0.87  & {}60.30\textpm 0.82 \tabularnewline
\hline 
\hline 
{}ResNet-18  & {}60.13\textpm 0.71  & {}78.79\textpm 0.54  & {}59.65\textpm 0.78  & {}76.75\textpm 0.73\tabularnewline
\hline 
{}ResNet-18+Dual TriNet  & \textbf{{}63.77}{}\textpm 0.62  & \textbf{{}80.53}{}\textpm 0.46  & \textbf{{}63.41}{}\textpm 0.64  & \textbf{{}78.43}{}\textpm 0.62\tabularnewline
\hline 
\end{tabular}
\par\end{centering}
\caption{\label{tab:cifar100}Results on Caltech-256 and CIFAR-100 datasets.
The ``{\small{}{}{}{}\textpm '' indicates $95\%$ confidence
intervals over tasks. }}

\end{table*}

\vspace{0.05in}

\noindent \textbf{(6) Even the semantic space inferred from the semantic
relationships of classes can also work well with our framework. }To
show this point, we again compare the results in Fig. \ref{fig:aug-layer-1}.
Particularly, we compute the similarity matrix of classes in \emph{mini}ImageNet
obtained using semantic word vectors. The SVD is employed to decompose
the similarity matrix and the left singular vectors of SVD are assumed
to span a new semantic space. Such a new space is hence utilized in
learning the dual TriNet. We employ the Semantic Gaussian (SG) to
augment the instance feature in the newly spanned space for one-shot
classification. The results are denoted ``SVD-G''. We report the
results of SVD-G augmentation in \emph{mini}ImageNet dataset in Fig.
\ref{fig:aug-layer-1}. We highlight several interesting observations.
(1) The results of SVD-G feature augmentation are still better than
those without any augmentation. (2) The accuracy of SVD-G is actually
slightly worse than that of SG, since the new spanned space is derived
from the original semantic word space. (3) There is almost no complementary
information in the augmented features between SVD-G and SG, still
partly due to the new space spanned from the semantic and the word
space. (4) The augmented features produced by SVD-G are also very
complementary to those from SN as shown in the results of Fig. \ref{fig:aug-layer-1}.
This is due to the fact that additional neighborhood vocabulary information
is not used in deriving the new semantic space. We have a similar
experimental conclusion on CUB-200 as shown in Tab. \ref{tab:CUB200}.

\begin{table*}
\centering{}%
\begin{tabular}{c|c|c|c|c|c|c|c|c|c|c}
\hline 
\multirow{2}{*}{{\small{}Method} } & \multirow{2}{*}{{\small{}Shots }} & \multirow{2}{*}{{\small{}R-18 }} & \multirow{2}{*}{{\small{}Layer} } & \multicolumn{7}{c}{{\small{}{}Data Augmentation}}\tabularnewline
\cline{5-11} 
 &  &  &  & \multicolumn{1}{c|}{{\small{}{}{}SN}} & \multicolumn{1}{c|}{{\small{}{}{}SG}} & \multicolumn{1}{c|}{{\small{}{}{}SD}} & \multicolumn{1}{c|}{{\small{}{}{}SN+SG}} & \multicolumn{1}{c|}{{\small{}{}{}SG+SD}} & \multicolumn{1}{c|}{{\small{}{}{}SN+SD}} & {\small{}{}{}SN+SG+SD }\tabularnewline
\hline 
\multirow{4}{*}{{\small{}KNN }} & \multirow{2}{*}{1 }  & \multirow{2}{*}{{\small{}{}64.30} }  & {\small{}{}{}S.}  & \multicolumn{1}{c|}{{\small{}{}{}65.12}} & \multicolumn{1}{c|}{{\small{}{}{}65.21}} & \multicolumn{1}{c|}{{\small{}{}{}65.38}} & \multicolumn{1}{c|}{{\small{}{}{}66.82}} & \multicolumn{1}{c|}{{\small{}{}{}65.50}} & \multicolumn{1}{c|}{{\small{}{}{}67.21}} & {\small{}{}{}67.23 }\tabularnewline
 &  &  & {\small{}{}{}M.}  & \multicolumn{1}{c|}{{\small{}{}{}65.58}} & \multicolumn{1}{c|}{{\small{}{}{}65.61}} & \multicolumn{1}{c|}{{\small{}{}{}65.78}} & \multicolumn{1}{c|}{{\small{}{}{}67.29}} & \multicolumn{1}{c|}{{\small{}{}{}65.77}} & \multicolumn{1}{c|}{{\small{}{}{}67.82}} & {\small{}{}{}67.91 }\tabularnewline
\cline{2-11} 
 & \multirow{2}{*}{5}  & \multirow{2}{*}{{\small{}{}77.66} }  & {\small{}{}{}S.}  & \multicolumn{1}{c|}{{\small{}{}{}78.34}} & \multicolumn{1}{c|}{{\small{}{}{}78.42}} & \multicolumn{1}{c|}{{\small{}{}{}78.62}} & \multicolumn{1}{c|}{{\small{}{}{}79.01}} & \multicolumn{1}{c|}{{\small{}{}{}78.66}} & \multicolumn{1}{c|}{{\small{}{}{}79.12}} & {\small{}{}{}79.36 }\tabularnewline
 &  &  & {\small{}{}{}M.}  & \multicolumn{1}{c|}{{\small{}{}{}79.01}} & \multicolumn{1}{c|}{{\small{}{}{}78.96}} & \multicolumn{1}{c|}{{\small{}{}{}79.04}} & \multicolumn{1}{c|}{{\small{}{}{}79.51}} & \multicolumn{1}{c|}{{\small{}{}{}79.09}} & \multicolumn{1}{c|}{{\small{}{}{}79.56}} & {\small{}{}{}79.71 }\tabularnewline
\hline 
\multirow{4}{*}{{\small{}SVR }} & \multirow{2}{*}{1}  & \multirow{2}{*}{{\small{}{}66.54} }  & {\small{}{}{}S.}  & \multicolumn{1}{c|}{{\small{}{}{}67.63}} & \multicolumn{1}{c|}{{\small{}{}{}67.49}} & \multicolumn{1}{c|}{{\small{}{}{}67.69}} & \multicolumn{1}{c|}{{\small{}{}{}68.23}} & \multicolumn{1}{c|}{{\small{}{}{}67.60}} & \multicolumn{1}{c|}{{\small{}{}{}68.41}} & {\small{}{}{}68.56 }\tabularnewline
 &  &  & {\small{}{}{}M.}  & \multicolumn{1}{c|}{{\small{}{}{}68.10}} & \multicolumn{1}{c|}{{\small{}{}{}68.03}} & \multicolumn{1}{c|}{{\small{}{}{}68.22}} & \multicolumn{1}{c|}{{\small{}{}{}68.71}} & \multicolumn{1}{c|}{{\small{}{}{}67.98}} & \multicolumn{1}{c|}{{\small{}{}{}68.89}} & {\small{}{}{}69.01 }\tabularnewline
\cline{2-11} 
 & \multirow{2}{*}{5 }  & \multirow{2}{*}{{\small{}{}82.38 }}  & {\small{}{}{}S.}  & \multicolumn{1}{c|}{{\small{}{}{}83.01}} & \multicolumn{1}{c|}{{\small{}{}{}83.07}} & \multicolumn{1}{c|}{{\small{}{}{}83.02}} & \multicolumn{1}{c|}{{\small{}{}{}83.59}} & \multicolumn{1}{c|}{{\small{}{}{}83.11}} & \multicolumn{1}{c|}{{\small{}{}{}83.42}} & {\small{}{}{}83.44 }\tabularnewline
 &  &  & {\small{}{}{}M.}  & \multicolumn{1}{c|}{{\small{}{}{}83.47}} & \multicolumn{1}{c|}{{\small{}{}{}83.51}} & \multicolumn{1}{c|}{{\small{}{}{}83.60}} & \multicolumn{1}{c|}{{\small{}{}{}83.82}} & \multicolumn{1}{c|}{{\small{}{}{}83.49}} & \multicolumn{1}{c|}{{\small{}{}{}83.99}} & {\small{}{}{}84.10 }\tabularnewline
\hline 
\multirow{4}{*}{{\small{}LR }} & \multirow{2}{*}{1}  & \multirow{2}{*}{{\small{}{}64.22} }  & {\small{}{}{}S.}  & \multicolumn{1}{c|}{{\small{}{}{}65.29}} & \multicolumn{1}{c|}{{\small{}{}{}65.33}} & \multicolumn{1}{c|}{{\small{}{}{}65.43}} & \multicolumn{1}{c|}{{\small{}{}{}66.59}} & \multicolumn{1}{c|}{{\small{}{}{}65.46}} & \multicolumn{1}{c|}{{\small{}{}{}66.89}} & {\small{}{}{}67.01 }\tabularnewline
 &  &  & {\small{}{}{}M.}  & \multicolumn{1}{c|}{{\small{}{}{}65.71}} & \multicolumn{1}{c|}{{\small{}{}{}65.92}} & \multicolumn{1}{c|}{{\small{}{}{}65.89}} & \multicolumn{1}{c|}{{\small{}{}{}67.12}} & \multicolumn{1}{c|}{{\small{}{}{}65.74}} & \multicolumn{1}{c|}{{\small{}{}{}67.63}} & {\small{}{}{}67.55 }\tabularnewline
\cline{2-11} 
 & \multirow{2}{*}{{\small{}{}5}}  & \multirow{2}{*}{{\small{}{}82.51} }  & {\small{}{}{}S.}  & \multicolumn{1}{c|}{{\small{}{}{}83.37}} & \multicolumn{1}{c|}{{\small{}{}{}83.31}} & \multicolumn{1}{c|}{{\small{}{}{}83.60}} & \multicolumn{1}{c|}{{\small{}{}{}83.61}} & \multicolumn{1}{c|}{{\small{}{}{}83.59}} & \multicolumn{1}{c|}{{\small{}{}{}83.62}} & {\small{}{}{}83.69 }\tabularnewline
 &  &  & {\small{}{}{}M.}  & \multicolumn{1}{c|}{{\small{}{}{}}\textbf{\small{}{}83.82}} & \multicolumn{1}{c|}{{\small{}{}{}}\textbf{\small{}{}83.83}} & \multicolumn{1}{c|}{{\small{}{}{}}\textbf{\small{}{}83.90}} & \multicolumn{1}{c|}{{\small{}{}{}}\textbf{\small{}{}84.21}} & \multicolumn{1}{c|}{{\small{}{}{}}\textbf{\small{}{}83.71}} & \multicolumn{1}{c|}{{\small{}{}{}}\textbf{\small{}{}84.23}} & {\small{}{}{}}\textbf{\small{}{}84.17 }\tabularnewline
\hline 
\end{tabular}{\tiny{}{}\caption{\label{tab:CUB200} \textbf{The classification accuracy of one-shot
learning on Caltech-UCSD Birds in 5-way. Note that: ``S.'' and ``M.''
indicates the single and multiple layers respectively. ``SD'' is
short for ``SVD-G''. ``R-18'' is short for ``ResNet-18''.}}
} 
\end{table*}

\subsection{Experimental results on Caltech-256 and CIFAR-100}

\begin{figure*}
\begin{centering}
\includegraphics[scale=0.3]{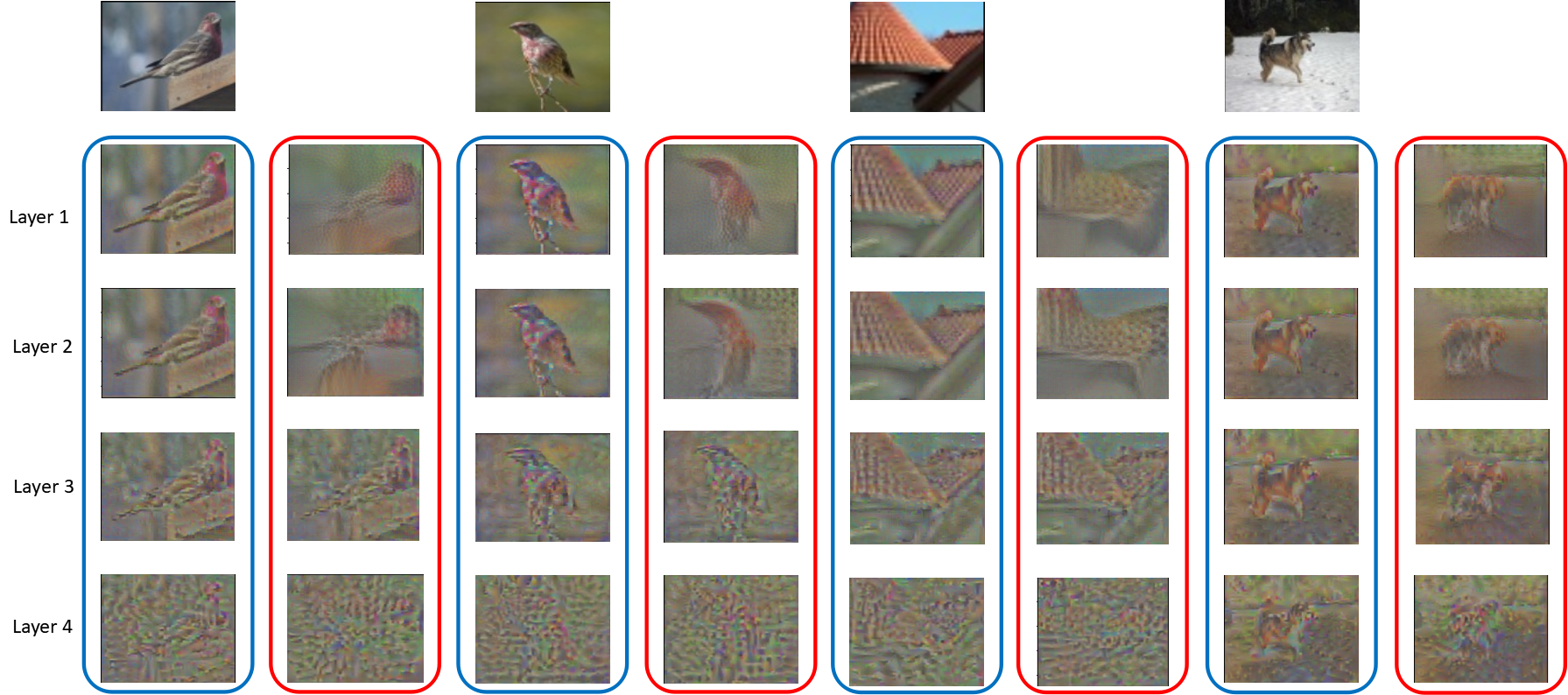} 
\par\end{centering}
\caption{\label{fig:Visualization-of-the}Visualization of the original and
augmented features.}
\vspace{-0.15in}
 
\end{figure*}

\noindent \textbf{Settings. }{On {}Caltech-256 and CIFAR-100 dataset
we also use the semantic word vector space. For one training instance,
we synthesize }16 sugmented features for SG and SN individually from
all four layers of ResNet-18. On these two datasets, the results of
competitors are implemented and reported in {\cite{DEML+Meta-SGD}}.
Our reported results are produced by using the augmented feature instances
of all layers, both by SG and SN. The SVM classifier is used as the
classification model.

\vspace{0.05in}

\noindent \textbf{Results. }The results on Caltech-256 and CIFAR-100
are illustrated in Tab. \ref{tab:cifar100}. We found that (1) our
method can still achieve the best performance as compared to the state-of-the-art
algorithms, thanks to the augmented feature instances obtained using
the proposed framework. (2) The ResNet-18 is still a very strong baseline;
and it can beat almost all the other baselines, except the DEML+Meta-SGD
which uses ResNet-50 as the baseline structure. (3) There is a clear
margin of improvement from using our augmented instance features over
using ResNet-18 only. This further validates the efficacy of the proposed
framework.

\section{Further analysis}

\begin{table*}
\centering{}%
\begin{tabular}{c|l|l|l|l|l|l|l|l}
\hline 
\multirow{2}{*}{Methods} & \multicolumn{2}{l|}{{}{}MiniImagenet} & \multicolumn{2}{l|}{{}{}CUB-200} & \multicolumn{2}{l|}{{}{}Caltech-256} & \multicolumn{2}{l}{{}{}CIFAR-100}\tabularnewline
\cline{2-9} 
 & {}{}1-shot  & {}{}5-shot  & {}{}1-shot  & {}{}5-shot  & {}{}1-shot  & {}{}5-shot  & {}{}1-shot  & {}{}5-shot \tabularnewline
\hline 
\hline 
{}{}ResNet-18  & {}{}52.73  & {}{}73.31  & {}{}66.54  & {}{}82.38  & {}{}60.13  & {}{}78.79  & {}{}59.65  & {}{}76.75 \tabularnewline
\hline 
{}{}ResNet-18+U-net  & {}{}56.41  & {}{}75.67  & {}{}68.32  & {}{}83.24  & {}{}61.54  & {}{}79.88  & {}{}62.32  & {}{}77.87 \tabularnewline
\hline 
{}{}ResNet-18+Auto-encoder  & {}{}56.80  & {}{}75.27  & {}{}68.56  & {}{}83.24  & {}{}62.41  & {}{}79.77  & {}{}61.76  & {}{}76.98 \tabularnewline
\hline 
{}Ours without encoder  & {}50.69  & {}70.79  & {}64.15  & {}80.06  & {}58.78  & {}76.45  & {}57.46  & {}75.12\tabularnewline
\hline 
{}Ours without decoder  & {}48.75  & {}65.12  & {}62.04  & {}78.16  & {}58.67  & {}76.45  & {}53.19  & {}68.74\tabularnewline
\hline 
\hline 
{}{}Ours  & \textbf{{}{}58.12}{}{}  & \textbf{{}{}76.92}{}  & \textbf{{}{}69.61}{}  & \textbf{{}{}84.10}{}{}  & \textbf{{}{}63.77}{}  & \textbf{{}{}80.53}{}  & \textbf{{}{}63.41}{}  & \textbf{{}{}78.43 }\tabularnewline
\hline 
\end{tabular}\caption{\label{tab:Alternative-augmentation-network}Results of using alternative
augmentation networks.Ours indicates ResNet-18+Dual-TriNet.}
\end{table*}

\subsection{Comparison with standard augmentation methods}

Besides our feature augmentation method, we also compare the standard
augmentation methods \cite{KrizhevskySH12} in one-shot learning setting.
These methods include cropping, rotation, flipping, and color transformations
of training images of one-shot classes. Furthermore, we also try the
methods of adding the Gaussian noise to the ResNet-18 features of
training instances of one-shot classes as shown in Tab. \ref{tab:miniimagenet}.
However, none of these methods can improve the classification accuracy
in one-shot learning. This is reasonable since the one-shot classes
have only very few training examples. This is somewhat expected: such
naive augmentation methods intrinsically just add noise/variance,
but do not introduce extra information to help one-shot classification.

\subsection{Dual TriNet structure}

We propose the dual TriNet structure which intrinsically is derived
from the encoder-decoder architecture. Thus we further analyze the
other alternative network structures for feature augmentation. In
particular, the alternative choices of augmentation network can be
the auto-encoder \cite{hinton2006science} of each layer or U-net
\cite{unet2015}. The results are compared in Tab. \ref{tab:Alternative-augmentation-network}.
We show that our dual TriNet can best explore the complementary information
of different layers, and hence our results are better than those without
augmentation (ResNet-18), with U-net augmentation ({ResNet-18+U-net)
and with auto-encoder augmentation (ResNet-18+Auto-encoder). This
validates that our dual TriNet can efficiently merge and exploit the
information of multiple layers for feature augmentation.} In addition,
we conduct experiments to prove that the encoder part and decoder
part is necessary. If we simply used the semantic vector of true label
$\mathbf{u}_{z_{i}}^{base}$instead of using encoder $g_{Enc}\left(\left\{ f_{l}\left(\mathbf{I}_{i}^{support}\right)\right\} \right)$,
the augmented samples actually hurt the performance. 
In the case where we do the classification in the semantic space,
effectively disabling the decoder, the performance drops by over 5\%.
This is because of the loss of information during the mapping from
visual space to semantic space, but in our approach, we keep original
information and have additional information from semantic space.

\subsection{Visualization }

Using the technique in \cite{2014arXiv1412.0035M}, we can visualize
the image that can generate the augmented features $\hat{f_{l}}\left(\mathbf{I}_{i}\right)=g\left(f_{l}\left(\mathbf{I}_{i}\right)\right)$
in ResNet-18. We first randomly generate an image $\mathbf{I}_{i_{0}}$.
Then we optimize $\mathbf{I}_{i_{0}}$ by reducing the distance betwen
$f_{l}\left(\mathbf{I}_{i_{0}}\right)$ and $\hat{f_{l}}\left(\mathbf{I}_{i}\right)$
(both are the output of ResNet-18): 
\begin{equation}
\mathbf{I}_{i_{0}}=\mathrm{\underset{\mathbf{I}_{i_{0}}}{\mathrm{argmin}}}\frac{1}{2}\left\Vert f_{l}\left(\mathbf{I}_{i_{0}}\right)-\hat{f_{l}}\left(\mathbf{I}_{i}\right)\right\Vert _{2}^{2}+\lambda\cdot R\left(\mathbf{I}_{i_{0}}\right)\label{eq:visualization}
\end{equation}
where $R\left(\cdot\right)$ is the Total Variation Regularizer for
image smoothness; $\lambda=1e-2$. When the difference is small enough,
$\mathbf{I}_{i_{0}}$ should be representation of the image that can
generate the corresponding augmented feature.

By using SN and the visualization algorithm above, we visualize the
original and augmented features in Fig. \ref{fig:Visualization-of-the}.
The top row shows the input images of two birds, one roof, and one
dog. The blue circles and red circles indicate the visualization of
original and augmented features of Layer 1 -- Layer 4 respectively.
The visualization of augmented features is similar, and yet different
from that of original image. For example, the first two columns show
that the visualization of augmented features actually slightly change
the head pose of the bird. In the last two columns, the augmented
features clearly visualize a dog which is similar have a different
appearance from the input image. This intuitively shows why our framework
works.

\section{Conclusions}

This work purposes an end-to-end framework for feature augmentation.
The proposed dual TriNet structure can efficiently and directly augment
multi-layer visual features to boost the few-shot classification.
We demonstrate our framework can efficiently solve the few-shot classification
on four datasets. We mainly evaluate on classification tasks; it is
also interesting and future work to extend augmented features to other
related tasks, such as one-shot image/video segmentation \cite{one_shot_seg17BMVC,one-shot-video17cvpr}.
Additionally, though dual TriNet is paired with ResNet-18 here, we
can easily extend it for other feature extractor networks, such as
ResNet-50. 

\bibliographystyle{splncs}
\bibliography{ref}

\end{document}